\documentclass[journal,transmag]{IEEEtran}
\usepackage{cite}
\usepackage{amssymb,latexsym,amsfonts}

\newtheorem{assumption}{Assumption}
\usepackage{pifont} 
\usepackage{enumitem}
\usepackage{multirow}

\newcommand{\indep}{\perp \!\!\! \perp}

\ifCLASSINFOpdf
  \usepackage[pdftex]{graphicx}

  \graphicspath{{../pdf/}{../jpeg/}}
  \DeclareGraphicsExtensions{.pdf,.jpeg,.png}
\else

\fi

\begin{document}

\title{Identification of Causal Relationship between Amyloid-$\beta$ Accumulation and Alzheimer’s Disease Progression via Counterfactual Inference}


\author{

\IEEEauthorblockN{Haixing Dai\textsuperscript{{$\dagger$}}\IEEEauthorrefmark{1},
Mengxuan Hu\textsuperscript{{$\dagger$}}\IEEEauthorrefmark{1},
Qing Li\textsuperscript{{$\dagger$}}\IEEEauthorrefmark{2,3},
Lu Zhang\IEEEauthorrefmark{4},
Lin Zhao\IEEEauthorrefmark{1},
Dajiang Zhu\IEEEauthorrefmark{4}
Ibai Diez\IEEEauthorrefmark{5},
Jorge Sepulcre\IEEEauthorrefmark{5}, \\
Fan Zhang\IEEEauthorrefmark{6},
Xingyu Gao\IEEEauthorrefmark{7},
Manhua Liu\IEEEauthorrefmark{8},
Quanzheng Li\IEEEauthorrefmark{5},
Sheng Li\IEEEauthorrefmark{9},
Tianming Liu\IEEEauthorrefmark{1},
and Xiang Li\IEEEauthorrefmark{5}
}
\IEEEauthorblockA{\IEEEauthorrefmark{1}School of Computing, University of Georgia, Athens, GA, USA}
\IEEEauthorblockA{\IEEEauthorrefmark{2}State Key Lab of Cognitive Neuroscience and Learning, Beijing Normal University, Beijing 100875, China}
\IEEEauthorblockA{\IEEEauthorrefmark{3}School of Artificial Intelligence, Beijing Normal University, Beijing 100875, China}
\IEEEauthorblockA{\IEEEauthorrefmark{4}Department of Computer Science and Engineering, The University of Texas at Arlington, Arlington 76019, USA}
\IEEEauthorblockA{\IEEEauthorrefmark{5}Department of Radiology, Massachusetts General Hospital and Harvard Medical School, Boston 02114, USA}
\IEEEauthorblockA{\IEEEauthorrefmark{6}Department of Radiology, Brigham and Women's Hospital and Harvard Medical School, Boston 02115, USA}
\IEEEauthorblockA{\IEEEauthorrefmark{7}School of Electronic Information and Electrical Engineering, Shanghai Jiao Tong University, Shanghai 200240, China}
\IEEEauthorblockA{\IEEEauthorrefmark{8}The  MoE  Key  Lab  of  Artificial  Intelligence,  AI Institute, Shanghai Jiao Tong University, Shanghai, 200240, China}
\IEEEauthorblockA{\IEEEauthorrefmark{9}School of Data Science, The University of Virginia, Charlottesville 22903, USA}
\thanks{
\textsuperscript{{$\dagger$}} These authors contributed equally to this paper.

Corresponding author: Xiang Li (email: xli60@mgh.harvard.edu).
}}

\IEEEtitleabstractindextext{%
\begin{abstract}
Alzheimer's disease (AD) is a neurodegenerative disorder that is beginning with amyloidosis, followed by neuronal loss and deterioration in structure, function, and cognition. The accumulation of amyloid-$\beta$ in the brain, measured through 18F-florbetapir (AV45) positron emission tomography (PET) imaging, has been widely used for early diagnosis of AD. However, the relationship between amyloid-$\beta$ accumulation and AD pathophysiology remains unclear, and causal inference approaches are needed to uncover how amyloid-$\beta$ levels can impact AD development. In this paper, we propose a graph varying coefficient neural network (GVCNet) for estimating the individual treatment effect with continuous treatment levels using a graph convolutional neural network. We highlight the potential of causal inference approaches, including GVCNet, for measuring the regional causal connections between amyloid-$\beta$ accumulation and AD pathophysiology, which may serve as a robust tool for early diagnosis and tailored care.
\end{abstract}

}

\maketitle

\IEEEdisplaynontitleabstractindextext

\IEEEpeerreviewmaketitle

\section{Introduction}
\label{intro}
\IEEEPARstart{T}{he}  differentiation of Alzheimer's disease (AD) from the prodromal stage of AD, which is the mild cognitive impairment (MCI), and normal control (NC) is an important project that interests many researchers making effort on~\cite{li2017multi,2022Somatic}. It is commonly recognized through studies that the progression of AD involves a series of gradually intensifying neuropathological occurrences. The process begins with amyloidosis, followed by neuronal loss and subsequent deterioration in the areas of structure, function, and cognition~\cite{naturemed2022}. As a non-invasive method that could measure the accumulation of amyloid in the brain, 18F-florbetapir (AV45) positron emission tomography (PET) imaging has been widely used for early diagnosis of AD~\cite{2022Tracer}. The use of florbetapir-PET imaging to characterize the deposition of amyloid-$\beta$ has shown to be of significant diagnostic value in identifying the onset of clinical impairment.

In recent years, there has been increasing research in counterfactual causal inference to estimate the treatment effect in various domains such as medicine~\cite{lv2021causal,yazdani2015causal,meilia2020review}, public health~\cite{rothman2005causation,glass2013causal,glymour2017evaluating}, and marketing~\cite{varian2016causal,hair2021data}. Especially, estimating the causal effect of continuous treatments is crucial. For example, in precision medicine, a common question is \emph{``What is the ideal medicine dosage to attain the best result?''}. Therefore, an average dose-response function (ADRF) that elucidates the causal relationship between the continuous treatment and the outcome becomes imperative.

\begin{figure}
    \centering
    \includegraphics[width=\linewidth]{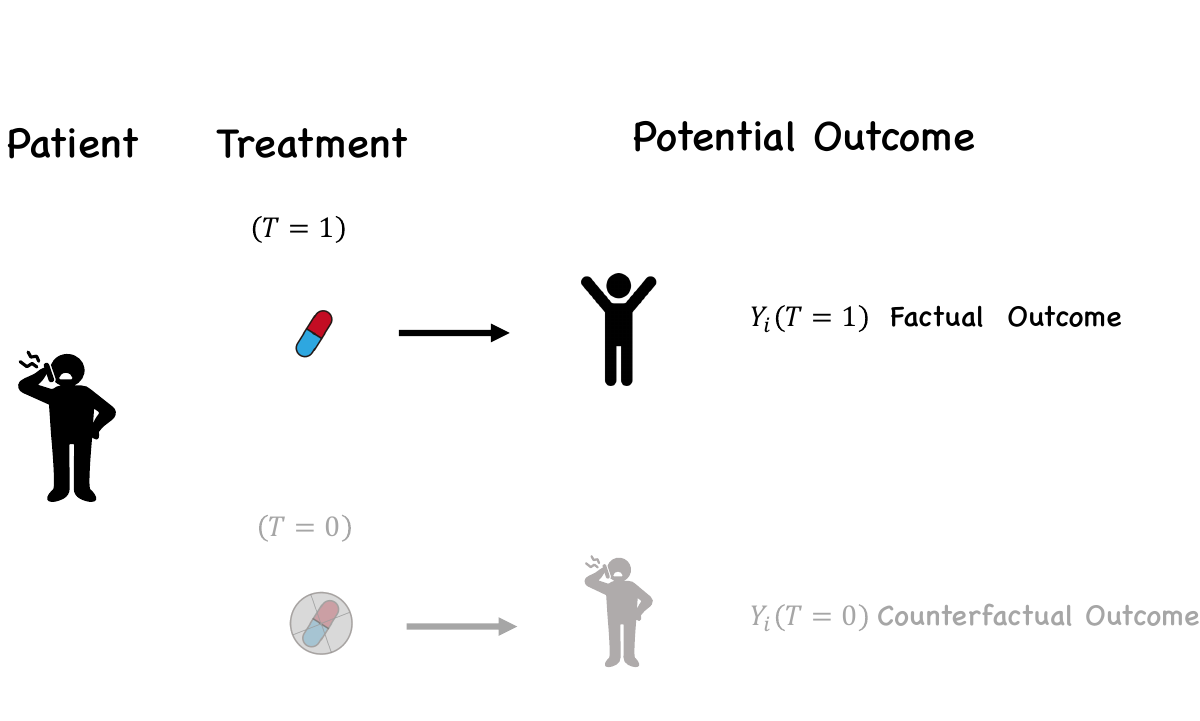}
    \caption{An Example of counterfactual problem: A patient with a headache who takes medicine and is cured. While the counterfactual scenario, i.e., the outcome had the patient not taken the medicine, is unobserved.}
    \label{fig:counterfactual}
\end{figure}

Estimating the counterfactual outcome presents a significant challenge in causal effect estimation, as it is inherently unobservable. To provide a clear definition, we use the binary treatment scenario ($T=1$ or $T=0$) for illustration. As depicted in Fig. \ref{fig:counterfactual}, let us consider a patient with a headache ($x_i$) who has the option to either take the medicine ($T=1$) or not take it ($T=0$). The potential outcomes corresponding to these two treatment choices would be being cured ($Y_i(T=1)$) or not being cured ($Y_i(T=0)$), respectively. The causal effect is defined as the difference between these two potential outcomes. However, given that a patient can only choose one treatment option, we can observe only one outcome (the observed outcome), while the other outcome that was not observed is considered the counterfactual outcome. Similarly, in the context of a continuous setting, estimating the counterfactual outcome remains a significant challenge.

Therefore, a variety of existing works on causal effect estimation focus on counterfactual estimation~\cite{morgan2015counterfactuals,johansson2016learning,hassanpour2019counterfactual} under the assumption of binary treatments or continuous treatments (ADRF estimation)~\cite{hirano2004propensity,schwab2020learning,nie2021vcnet,bica2020estimating,zhang2022exploring}. 
Especially, in the context of continuous treatments, the generalized propensity score (GPS), proposed by Hirano and Imbens~\cite{hirano2004propensity}, is a traditional approach to estimate ADRF with counterfactual outcomes. Moreover, as machine learning has gained increasing attention due to its extraordinary ability to solve complex problems, many existing works use machine learning techniques to address the problem. Schwab et al.~\cite{schwab2020learning} proposed DRNet to split a continuous treatment into several intervals and built separate prediction heads for them on the latent representation of input. Nie et al.~\cite{nie2021vcnet} adopted varying coefficient structure to explicitly incorporate continuous treatments as a variable for the parameters of the model, preserving the continuity of ADRF. Other methods, such as GAN~\cite{bica2020estimating} and transformer~\cite{zhang2022exploring}, have also been proposed.

In this work, we propose a novel model, the Graph Varying Coefficient Neural Network (GVCNet), for measuring the regional causal associations between amyloid-$\beta$ accumulation and AD pathophysiology. Specifically, by comparing our model with the most advanced model, VCNet, we demonstrate that our model achieves better performance in AD classification. Moreover, we adopt K-Means clustering to group the generated average dose-response function (ADRF) curves from each region of interest (ROI) and then map them onto the cortical surface to identify the amyloid-$\beta$ positive regions.

The main contributions of this work are summarized as follows:

1. To the best of our knowledge, this is the early attempt to utilize the brain structural topology as the graph to measure the regional causal associations between amyloid-$\beta$ accumulation and AD pathophysiology. Consistent experimental results on AD public dataset not only demonstrate the effectiveness and robustness of the proposed framework, but also support this hypothesis: the AD pathophysiology is deeply associated with amyloid-$\beta$ accumulation, no matter with which kind of topology graph.
2. Compared with the most advanced approach (i.e., VCNet), the proposed GVCNet experimentally obtains a higher diagnosis accuracy, suggesting that the good performance could be achieved with graph topology. As such our framework, such attempt extends the applications of graph-based algorithms on brain imaging analysis and provides a new insight into the causal inference that combines the phenotype, structural and functional data.
3. Our work demonstrates clearly that there are four brain regions (i.e., pre- \& post- central gyrus among cortical area, left \& right pallidum among subcortical area) can be as the key ROIs for AD diagnosis. With the quantitative experimental results, with such ROIs, the diagnosis accuracy is better than with the whole brain information.

\section{Related Work}
\subsection{Counterfactual Outcome Estimation}

 The definition of counterfactual outcome is typically framed using the potential outcome framework~\cite{rubin1974estimating}. To provide a clear definition, we illustrate with the use of binary treatments, which can be extended to multiple treatments by comparing their potential outcomes. Each individual $x_i$ has two potential outcomes: $Y_i(T=1)$ and $Y_i(T=0)$, corresponding to the two possible treatments ($T=1$ or $T=0$). Since an individual can only receive one of the two treatments in observational data, only one potential outcome can be observed (observed outcome), while the remaining unobserved outcome is referred to as the counterfactual outcome. Hence, the major challenge in estimating Individual Treatment Effect (ITE) lies in inferring counterfactual outcomes. Once the counterfactual outcomes are obtained, ITE can be calculated as the difference between the two potential outcomes:
\begin{equation}
ITE_i= Y_i(T=1)- Y_i(T=0).
\end{equation}

Many existing approaches have been proposed to estimate the counterfactual outcomes, such as conditional outcome modeling that trains two separate models to predict outcomes for the treatment group and control group and use the predicted value to fill the unobserved counterfactual outcomes. In addition, tree-based and forest-based methods are widely used to estimate ITE~\cite{chipman2010bart,hansen2008prognostic,wager2018estimation}.
Additionally, matching methods~\cite{morgan2015counterfactuals,stuart2010matching}, stratification mathods~\cite{yao2022concept}, deep representation methods~\cite{hassanpour2019counterfactual,yao2022concept} have been proposed to address the problem as well.

\subsection{Continuous Treatment Effect Estimation}
Continuous treatments are of great practical importance in many fields, such as precision medical. Typically, the objective of continuous treatment effect estimation is to estimate the average dose-response function (ADRF), which demonstrates the relationship between the specific continuous treatment and the outcome. 
Although recent works utilized the representation learning methods for ITE estimation~\cite{johansson2016learning,shalit2017estimating,chu2020matching,yao2018representation}, most of the existing works are under the assumption of binary treatments, which cannot be easily extended to continuous treatment due to their unique model design. 
To address this issue, Schwab et al.~\cite{schwab2020learning} extended the TARNet~\cite{shalit2017estimating} and proposed Dose Response networks (DRNet), which divided the continuous dosage into several equally-sized dosage stratus, and assigned one prediction head for each strata. To further achieve the continuity of ADRF, Nie et al.,~\cite{nie2021vcnet} proposed a varying-coefficient neural network (VCNet). Instead of the multi-head design, it used a varying coefficient prediction head whose weights are continuous functions of treatment $t$, which improved the previous methods by preserving a continuous ADRF and enhancing the expressiveness of the model. Hence, in this paper, we adopt it as part of the model to estimate the effect of each Regions of Interest (ROI) of the brain on Alzheimer's disease. 

\subsection{Traditional Correlation-based PET Image Analysis Methods}

The correlation-based methods on PET images analysis could be used in many clinical applications, such as tumor detection and brain disorder diagnosis.  An et al. used canonical correlation analysis-based scheme to estimate a standard-dose PET  image from a low-dose one in order to reduce the risk of radiation exposure and preserve image quality \cite{7468470}. Landau et al. used the traditional corrlation method to compare the retention of the 11-C radiotracer Pittsburgh Compound B and that of two 18-F amyloid radiotracers (florbetapir and flutemetamol) \cite{Landau}. Zhu et al. used the cannoical representation to consider the correlations relationship between features of  PET and other different brain neuroimage modalities \cite{10.1007/978-3-319-10470-6_21}. Li et al. used sparse inverse covariance estimation to reveal the relationship between PET and structural magnetic resonance imaging (sMRI) \cite{Li2018}. 

And for the AD diagnosis, it has been suggested that brain regions such as the posterior cingulate and lateral temporal cortices are affected more in AD than the NC, with the florbetapir-PET~\cite{2012Using}. Some researches on florbetapir-PET imaging have revealed that neurodegeneration does not influence the level of amyloid-$\beta$ accumulation. Instead, amyloid-$\beta$ pathophysiology is considered a biologically independent process and may play a "catalyst" role in neurodegeneration~\cite{2014Rates}. There have also been many theories that highlight the amyloid-$\beta$ pathologies as the main driving forces behind disease progression and cognitive decline. In order to characterize the relationship between the amyloid-$\beta$ accumulation and AD pathophysiology, the counterfactual causal inference method will be a useful tool to uncover how the patterns of causality or significant changes in regional or temporal amyloid-$\beta$ levels can impact the development of AD over time. 
\begin{figure*}
    \centering
    \includegraphics[width=16cm]{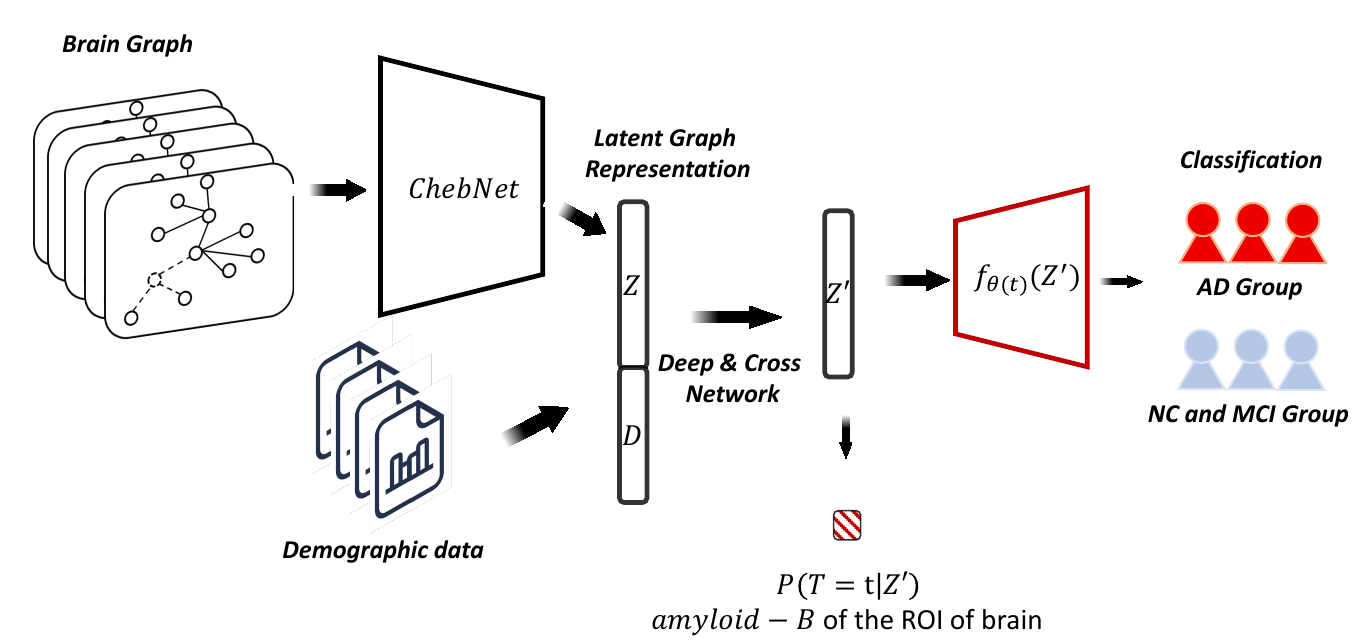}
    \caption{The framework of GVCNet for AD classification and individual treatment effect estimation. (a) we utilize ChebNet for feature embedding and then integrate treatment in the following dynamic fully connected layer for AD classification task. (b) We employee KMeans cluster algorithm to cluster the individual ADRFs into 3 groups: a$\beta$-positive (up), a$\beta$-negative (down) and a$\beta$-neutral and mapping these groups on the brain. }
    \label{fig:framework}
\end{figure*}

\subsection{Graph Neural Network}
Deep learning has revolutionized many machine learning tasks, but challenges arise when data is represented as graphs. The basic idea behind GNNs is to iteratively update the feature vectors of each node by aggregating the feature vectors of its neighboring nodes.

The update rule for a GNN can be formalized as follows:

\begin{equation}
    h^{l+1}_i = \sigma(\textbf{a}_i^l W^l), \textbf{a}_i^l = g^l(h_i^l, \{h_u^l: u \in \mathcal{N}(i)\}),
\end{equation}


where $h_i^{(l+1)}$ is the feature vector of node $i$ at layer $l+1$, $\mathcal{N}(i)$ is the set of neighboring nodes of $i$, $g^l$ is the aggregation function at latyer $l$, and $W^{(l)}$ is a learnable weight matrix at layer $l$. The function $\sigma$ is a non-linear activation function, such as the ReLU function. Graph convolutional networks (GCNs) extend convolutional neural networks~\cite{lecun1995convolutional} to the graph domain, allowing for meaningful feature extraction. GCNs have been applied in various fields, including node classification~\cite{wang2017mgae}, link prediction~\cite{zhang2018link}, and graph generation~\cite{kawamoto2018mean}. Initial work on GCNs was proposed by~\cite{gori2005new} in 2013, followed by the seminal paper by~\cite{kipf2016semi} in 2017. Since then, many extensions and improvements to GCNs have been proposed, including Graph Attention Networks (GATs)~\cite{velivckovic2017graph} and GraphSAGE~\cite{hamilton2017inductive}. Researchers have also studied different graph convolutional layers, such as Message Passing Neural Networks (MPNNs)~\cite{gilmer2017neural} and Convolutional Graph Neural Networks (ConvGNNs)~\cite{schlichtkrull2018modeling}. Overall, GCNs have shown great potential in graph representation learning and have the potential to revolutionize many applications where data is represented in the form of graphs.

\section{Methodology}
\subsection{Problem Setting}
VCNet is one of the advanced methods for ADRF estimation, typically it can generate continuous ADRF and provide promising counterfactual estimation.
Hence, in this study, we adopt this model to estimate the effect between the amyloid-$\beta$ level and the probability of gaining AD. Typically, we treat the amyloid-$\beta$ in a specific brain region as the treatment $T$ and whether the subject gains AD as the outcome $Y$. 

In our study, we used the Harvard-Oxford Atlas (HOA) to divide the entire brain into 69 regions. Since the some regions for tau imaging is not a target binding region, we excluded the following regions: left cerebral white matter, left cerebral cortex, left lateral ventrical, right cerebral white matter, right cerebral cortex, right lateral ventricle and brain-stem. For the rest of 62 regions, we treated one region as the treatment and used the other regions as covariates (X) to train a separate model for each setting. We iterated this process 62 times to obtain the causal effect and accuracy estimates for each region. To capture more information, we used graph structures of the whole brain denoted as $\mathcal{G} = (\mathcal{V},\mathcal{E},\mathcal{X})$, where each graph contains 62 nodes representing 62 ROIs, $\mathcal{V}$ represents the node set and $\mathcal{E}$ represents the edge set. Let $X \in R^{N \times F}$ be the input feature matrix, where each row corresponds to a node and each column corresponds to a feature. 
To estimate the causal effect of one ROI, we removed the corresponding node and all edges related to it and used the rest of the graph as input (61 nodes). Finally, we used the amyloid-$\beta$ value as the treatment variable $T$ for the VCNet analysis. In our work, we follow three fundamental assumptions for identifying ADRF:

\begin{assumption}
\textbf{Stable Unit Treatment Value Assumption (SUTVA)}: 
There are no unit interactions, and there is only one version of each treatment, which means that various levels or doses of a specific treatment are considered as separate treatments.

\end{assumption}

\begin{assumption}
\textbf{Positivity}: Every unit should have non-zero probability of being assigned to every treatment group. Formally, $P(T=t|X=x)\neq 0, \forall t\in \mathcal{T}, \forall x\in X$.

\end{assumption}

\begin{assumption}
\textbf{Ignorability}: Given covariates $x$, 
all potential outcomes $\{Y(T=t)\}_{t\in \mathcal{T}}$ are independent of the treatment assignment, implying that there are no unobserved confounders. Mathematically, $\{Y(T=t)\}_{t\in \mathcal{T}} \indep T | X$.
\end{assumption}

\begin{table*}[]
\centering
\caption{Description of the ADNI1 and ADNI2 datasets used in this work. N is the number of participants in each group; \textit{p}-value is calculated based on ANOVA}
\label{TAB:data}
\begin{tabular}{ccccccccccc}
\hline  
                       & Groups & N   & Age        & \textit{p}-value        & Sex(M/F) & \textit{p}-value        & MMSE Score & \textit{p}-value & CDR Score  & \textit{p}-value \\ \hline  
\multirow{3}{*}{ADNI1} & NC     & 100 & 75.83 4.71 & 0.4416 & 61/39    & 0.3923 & 28.94 1.12 & 0 & 0.0 0.0    & 0 \\
                       & MCI    & 205 & 74.98 7.23 &          & 136/69   &          & 27.18 1.69 &   & 0.49 0.03  &   \\
                       & AD     & 92  & 75.87 7.33 &          & 54/38    &          & 23.48 2.11 &   & 0.81 0.244 &   \\
\multirow{3}{*}{ADNI2} & NC     & 159 & 76.63 6.33 & 0.1436   & 77/82    & 0.2099 & 28.63 1.69 & 0 & 0.09 0.21  & 0 \\
                       & MCI    & 143 & 75.04 7.43 &          & 74/69    &          & 24.71 4.50 &   & 0.68 0.53  &   \\
                       & AD     & 106 & 76.29 7.95 &          & 63/43    &          & 20.02 4.60 &   & 1.06 0.48  &  \\ \hline  
\end{tabular}
\end{table*}

\begin{table*}[h]
\centering
\caption{Evaluation on GVCNet. $\star$ means the demographic feature is selected.}
\label{TAB:cross}
\begin{tabular}{cccccccc}
\hline   
Dataset  & Graph & Age & Sex & MMSE & CDR & Accuracy (\%)   \\ \hline   
ADNI1+ADNI2      & Corr   &     &     &      &     & 0.8296 $\pm$ 0.0020 \\
ADNI1+ADNI2     & Corr   & $\star$  & $\star$   &      &     & 0.8675 $\pm$ 0.0018 \\
ADNI1+ADNI2     & Corr  & $\star$   & $\star$   & $\star$    & $\star$   & 0.8868 $\pm$ 0.0027 \\ 
ADNI1+ADNI2      & DTI   &     &     &      &     & 0.8698 $\pm$ 0.0019 \\
ADNI1+ADNI2     & DTI   & $\star$  & $\star$   &      &     & 0.8689 $\pm$ 0.0018 \\
ADNI1+ADNI2     & DTI   & $\star$   & $\star$   & $\star$    & $\star$   & 0.8872 $\pm$ 0.0022 \\ \hline
\end{tabular}
\end{table*}

\subsection{GVCNet}
In our proposed GVCNet framework, as illustrated in Figure \ref{fig:framework}, there are three main components: ChebNet~\cite{defferrard2016convolutional}, Deep\&Cross Network~\cite{wang2017deep}, and VCNet~\cite{nie2021vcnet}. These components work together to estimate the Average Treatment Effect (ATE) using graph-structured data and demographic information.

The ChebNet component takes advantage of the graph structure of the data and utilizes this graph structure to generate features or representations that capture the underlying relationships between entities. 

The Deep\&Cross Network component incorporates demographic data into the framework. The Deep\&Cross Network module utilizes these demographic features to learn complex interactions between them, capturing both low-order and high-order feature interactions. This helps to capture additional information beyond what can be learned solely from the graph-structured data.

The resulting latent representation, denoted as $Z^{\prime}$, which is a combination of features from ChebNet and Deep\&Cross Network, is then fed into the VCNet component. VCNet infers the treatment distribution from $Z^{\prime}$ to ensure that it contains sufficient information for accurate ADRF estimation. Finally, the ADRF is estimated based on $t$ and $Z^{\prime}$.


\subsection{ChebNet}
In this paper, to preserve the topological information of PET data. We introduce the Chebyshev neural network (ChebNet)~\cite{defferrard2016convolutional} to replace the first two fully connected layers in VCNet. ChebNet uses Chebyshev polynomials to approximate the graph Laplacian filter, which is a commonly used filter in GCNs. Chebyshev polynomials are a sequence of orthogonal polynomials that can be used to approximate any smooth function on a given interval, and can be efficiently computed using recursive formulas.
The equation of first ChebNet is as follows:
\begin{equation}
    f_{\mathrm{out}}(\mathcal{L}, \mathbf{X})=\sigma\left(\sum_{k=0}^{K-1} \Theta_{k} T_{k}(\tilde{\mathcal{L}}) \mathbf{X}\right)    
\end{equation}
where $\mathbf{X} \in \mathbb{R}^{N \times F}$ is the input matrix of $N$ nodes, each with $F$ features, $\mathcal{L}$ is the graph Laplacian, and $\tilde{\mathcal{L}}$ is the normalized Laplacian defined as $\tilde{\mathcal{L}} = 2\mathcal{L}/\lambda_{\max} - I_N$, where $\lambda_{\max}$ is the largest eigenvalue of $\mathcal{L}$.  $T_k(\cdot)$ are Chebyshev polynomials of order $k$ and $\Theta_{k}$ are the learnable filter coefficients for the $k$-th Chebyshev polynomial. Finally, $\sigma(\cdot)$ is a non-linear activation function such as ReLU or sigmoid that is applied element-wise to the output of the ChebNet. And the binary cross-entropy loss function is utilized to quantify the dissimilarity between the predicted probability of the positive class and its true probability in binary classification tasks.

\subsection{Deep \& Cross Network}
The Deep \& Cross Network (DCN)~\cite{wang2017deep} is utilized to combine demographic data with topological information from PET data. Instead of conducting task-specific feature engineering, the DCN is capable of automatically learning the interactions between features that contribute to the task. Although deep neural networks (DNNs) are capable of extracting feature interactions, they generate these interactions in an implicit way, require more parameters, and may fail to learn some feature interactions efficiently.

The DCN uses an embedding and a stack layer to embed sparse features in the input into dense embedding vectors $x_{embed,k}^T$ to reduce the dimension. These vectors are then stacked with normalized dense features $x_{dense}^T$ in the input as a single vector $x_0=[x_{embed,1}^T,...,x_{embed,k}^T,x_{dense}^T ]$. A cross network and a deep network are adopted to further process this vector in parallel. The hallmark of the paper is the cross network, which applies explicit and efficient feature crossing as shown below:

\begin{equation}
\label{dcn}
x_{l+1} =x_0x_l^Tw_l+b_l+x_l
\end{equation}

Here, $x_l$ denotes the output of the $l$-th cross layer, and $w_l$ and $b_l$ represent the weight and bias of the $l$-th cross layer, respectively. The equation demonstrates that the degree of feature interactions grows with the depth of the layer. For example, the highest polynomial degree of $x_0$ of an $l$-layer cross network is $l+1$. Additionally, the interactions in the deep layer depend on the interactions in shallow layers.

In addition to the cross network, a fully-connected feed forward neural network is used to process $x_0$ simultaneously. The outputs of the cross network and the deep network are concatenated and fed into a standard logit layer to conduct the final prediction by the combination layer.

\subsection{VCNet}
Despite the prior endeavours on ITE estimation, most of the work are focused on binary treatment settings and fail to extend to continuous treatment easily. Although some papers propose to estimate the continuous treatment by splitting the range of treatment into severel intervals and use one prediction network for each interval, the continuity of ADRF is still an open issue. To address these issues, VCNet is proposed by \cite{nie2021vcnet}, which is capable of estimating continuous treatment effect and maintaining the continuity of ADRF simultaneously. 

A fully connected feedforward neural network is trained to extract latent representation $z$ from input $x$. To guarantee $z$ encode useful features, $z$ is used to estimate the conditional density of the corresponding treatment $\mathbb{P}(t|z)$ through a conditional probability estimating head. Specifically, $\mathbb{P}(t|z)$ is estimated based on the $(B + 1)$ equally divided grid points of treatment and the conditional density for the remaining t-values is computed using linear interpolation. After obtaining the $z$ containing valuable information, a varying coefficient neural network $f_{\theta(t)}(z)$ is adopted to predict the causal effect of $t$ on the outcome $y_{i,t}$ based on $z$ and the corresponding $t$, where the network parameters are a function of treatment $f_{\theta(t)}$ instead of fixed parameters. Typically, the B-spline is used to model $\theta(t)$:
\begin{equation}
    \theta(t)=[\sum_{l=1}^{L}a_{1,l}
    \varphi^{NN}_{l}(t),
    \cdots
    ,\sum_{l=1}^{L}a_{d_{\theta(t)},l}\varphi^{NN}_{l}(t)]^T \in \mathbb{R}^{d(\theta)},
\end{equation}

$\varphi^{NN}_{l}(t)$ denotes the spline basis of the treatment and $a_{1,l}$ are the coefficients to be learned; $d(\theta)$ is the dimension of $\theta(t)$.
By utilizing the varying coefficient neural network, the influence of the treatment effect $t$ on the outcome is integrated via the parameters of the outcome prediction network, thereby preventing any loss of treatment information. Additionally, the incorporation of $t$ in this manner allows for the attainment of a continuous ADRF.

\section{Experiment}
\subsection{Dataset}
In this paper, we conducted an evaluation of their proposed algorithm using two subsets of data from the Alzheimer's Disease Neuroimaging Initiative (ADNI) database (adni.loni.usc.edu), specifically ADNI-1 and ADNI-2, as well as the entire dataset. The subjects were divided into three categories, consisting of AD, NC, and MCI, as shown in Table \ref{TAB:data}. In this paper, we take AD as the AD group (298 subjects) and NC+MCI as the non-AD group (607 subjects). All florbetapir-PET images were co-registered with each individual’s sMRI and subsequently warped to the cohort-specific DARTEL template. And all subject has demographic features: age, sex, CDR score and MMSE score. 

All sMRI and florbetapir-PET images in this study are pre-processed by FMRIB Software Library (FSL) 6.0.3 (https://fsl.fmrib.ox.ac.uk/). The brain extraction step is based on the BET algorithm firstly\cite{hbm2002}. And the skull is stripped from the source image sapce. Secondly, the sMRI images are aligned to Montreal Neurological Institute T1 standard template space (MNI152) with the FLIRT linear registration algorithm\cite{2002Improved}, which can save computational time during the application stage. All florbetapir-PET images were co-registered with each individual's sMRI and subsequently warped to the cohort-specific DARTEL template. More specifically, after registration, the sMRI and florbetapir-PET images are cropped to the size of 152 × 188 × 152 by removing the voxels of zero values in the periphery of brain. Then, all the images are downsampled to the size of  76 × 94 × 76 that to reduce the computational complexity. And all subject has demographic features: age, sex, CDR score and MMSE score. 

In order to generate the structural connectivity matrix between different cortical regions, we also used the T1w and diffusion MRI (dMRI) provided in the ADNI database. T1-weighted images were acquired using a 3D sagittal MPRAGE volumetric sequence with TE = 3.0 ms; TI = 900.0 ms; TR = 2300.0 ms; flip angle = 9°; matrix size = 176 × 240 × 256; voxel size = 1.2 × 1.1 × 1.1 mm3.  dMRI was acquired with a spin-echo planar imaging (EPI) sequence. 48 noncollinear gradient directions were acquired with a b-value of 1,000 s/mm2. 7 additional volumes were acquired without diffusion weighting (b-value = 0 s/mm2). Other parameters of dMRI were as follows: TE = 56.0 ms; TR = 7200.0 ms; flip angle = 90°; matrix size = 116 × 116 × 80; isotropic voxel size = 2 × 2 × 2 mm3. A subset of 20 subjects was used for generating a group-wise connectivity matrix. For each subject, whole brain tractography was computed using the dMRI data, with the Unscented Kalman Filter (UKF) tractography method \cite{wan2000unscented,wan2001unscented} provided in the SlicerDMRI \cite{norton2017slicerdmri,zhang2020slicerdmri} software. Structural T1w imaging data was processed using FreeSurfer (version 6.0, https://surfer.nmr.mgh.harvard.edu/), and cortical regions were parcellated with the Desikan-Killiany Atlas \cite{alexander2019desikan}. Co-registration between the T1-weighted and dMRI data was performed using FSL \cite{jenkinson2012fsl}. Then, for each pair of cortical regions, streamlines that end in the two regions were extracted and the number of streamlines were computed, followed by the creation of the subject-specific connectivity matrix. For the group-wise connectivity matrix, the mean number of streamlines across the 20 subjects was recorded.

In the trainning process, We randomly split the dataset into a training set (633 subjects) and a testing set (272 subjects). The proposed model was tested on the testing set to calculate the classification accuracy and generate average dose-response function curves (ADRFs) for each ROI.

\subsection{Experiment Setting}

In GVCNet, we designate each one of the 62 ROIs as the treatment and use the other ROIs as patient features. The average amyloid-$\beta$ level serves as the signal for each ROI. We construct the input graph by defining the ROIs as nodes $V$ and the DTI structure among the ROIs as edges $E$. For the sturctural connectivity matrix, we have two alternative cunstructing options as follows: one is to use the Pearson correlation value among the ROIs' T1-weighted values to construct the structural correlation
graph (which is called the Corr graph in this paper to make it simplified); the other is to use the smoothed white fibers among the ROIs based on the 20 subjects (which is called DTI graph). Then treat the graph embedding and demographic data as input of the deep and cross network. Finally, feed the treatment and calculate the counter-factor with our GVCNet. For the hyper-parameters, we set the learning rate to 1e-4 and $\beta$ to 0.5. During model training, all networks were trained for 600 epochs. Our model is trained using Adam \cite{kingma2014adam} with momentum as 0.9.

\begin{table}[h]
\centering
\caption{Evaluation on GVCNet and VCNet on ADNI1+ADNI2}
\label{table2}
\begin{tabular}{lc}
\\ \hline
            &   Average Accuracy \\ \hline
VCNet       &  0.8401 $\pm$ 0.0048              \\
Graph-VCNet &   \textbf{0.8872 $\pm$ 0.0022}     \\ \hline       
\end{tabular}
\end{table}

\begin{figure*}
    \centering
    \includegraphics[width=18cm]{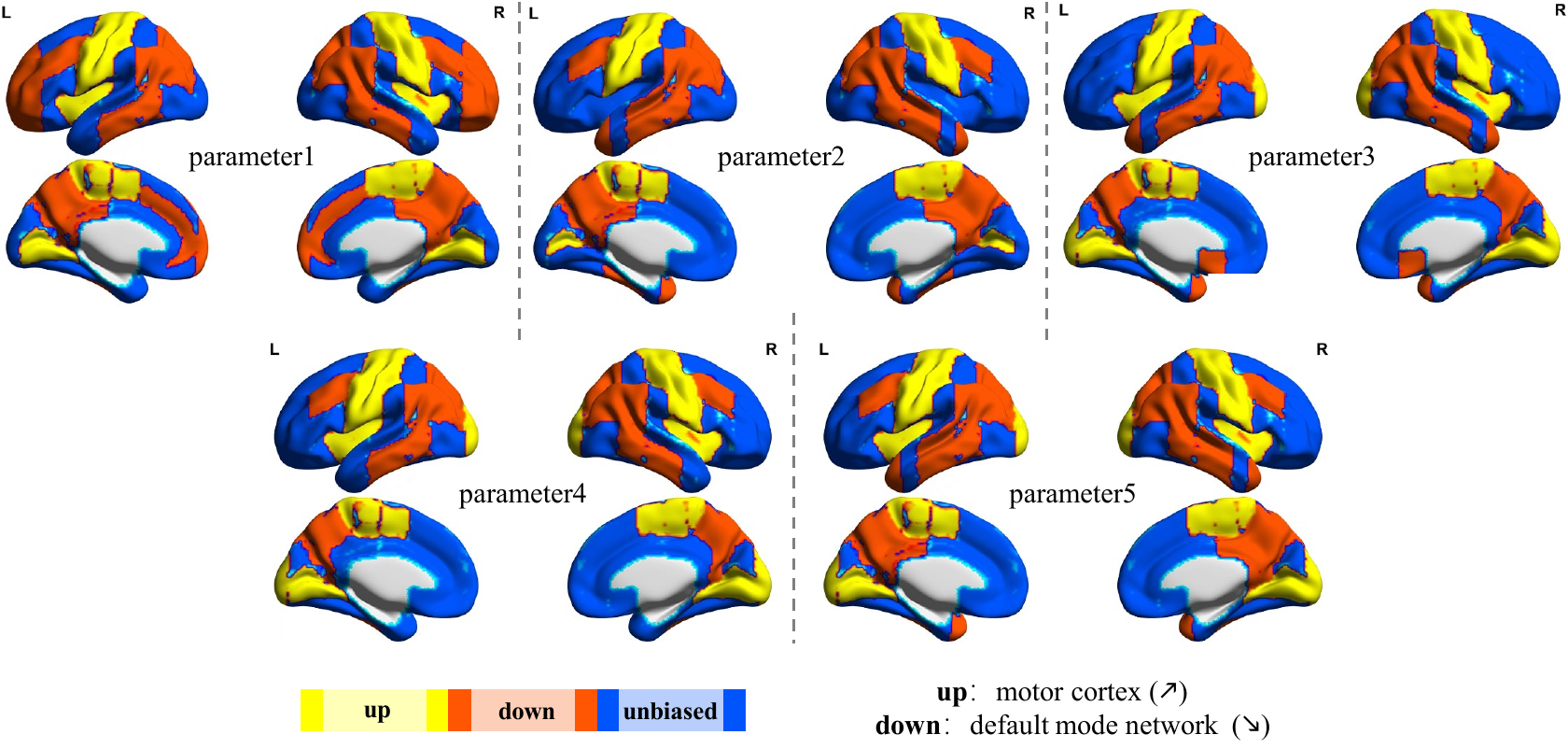}
    \caption{The cortical curve trends clustered by k-means. }
    \label{fig:cort}
\end{figure*}

\begin{figure*}
    \centering
    \includegraphics[width=18cm]{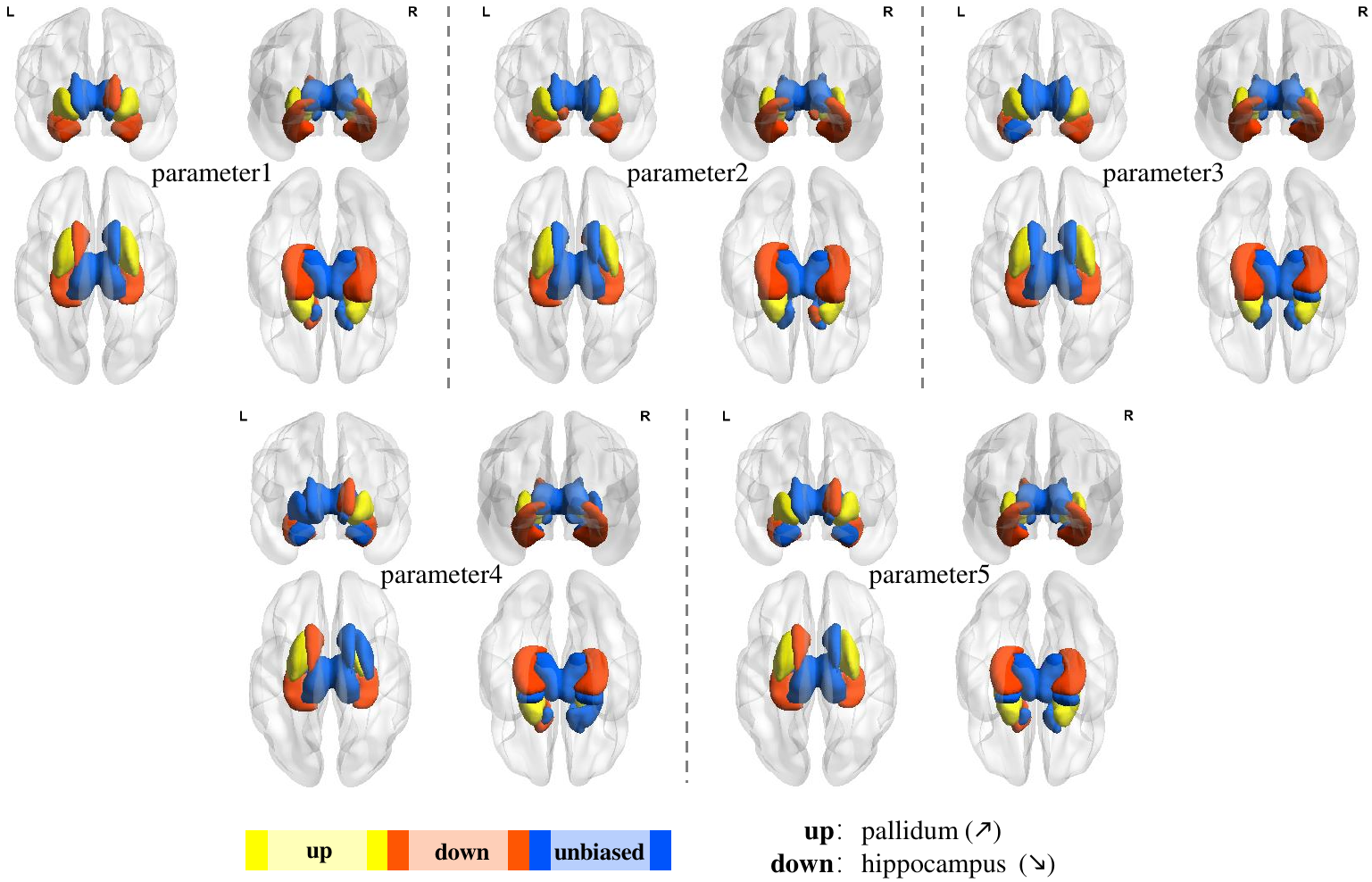}
    \caption{The subcortical curve trends clustered by k-means. }
    \label{fig:sub}
\end{figure*}

\subsection{Prediction Performance}
First, we compare our model, GVCNet with the baselien model, VCNet. As shown in Table \ref{table2}, the prediction performance of our model is around 88.72\%, which is 4.7\% higher than VCNet. 
In Table \ref{TAB:cross}, we evaluate the model's performance by the accuracy percentage. The table presents the evaluation results of the GVCNet model on different datasets, using different types of graphs, and considering different demographic factors. 

The first three rows present the evaluation results on the combined ADNI1+ADNI2 dataset, using Corr graphs and again different combinations of demographic factors. The model achieves an average accuracy of 0.8296 when no demographic features are selected, an average accuracy of 0.8675 when age and sex are used, and an average accuracy of 0.8868 when all the demographic features are selected. 

The last three rows present the evaluation results on the combined ADNI1+ADNI2 dataset, using DTI graphs and again different combinations of demographic factors. The model achieves an accuracy of 0.8698 when no features are selected, an accuracy of 0.8689 when age and sex features are considered, and an accuracy of 0.8872 when all the features are selected.
By comparing the last 6 rows, we can see that using DTI as the graph structure is slightly better than using the correlation graph between the ROIs as the graph structure.

\subsection{ADRF Curve Analysis}
Based on the patterns of the estimated ADRF of each region and the premise that different parts of the brain may play different roles during the normal/abnormal aging process, we use KMeans clustering method to cluster the ADRF curves from each region into three groups: upward(up, a$\beta$ positively respond to the treatment), downward(down, a$\beta$ negatively respond to the treatment) and unbiased, based on their trend of relationship with AD probability. Brain regions within each cluster were visualized onto the cortex and subcortex mappings in Fig. \ref{fig:cort} and Fig. \ref{fig:sub}. It can be found that there exist strong causal relationships between the AD progression and the PET signal level in the precentral/postcentral gyrus (cortical) and left/right pallidum (subcortical), indicating the potentially important role of these regions in modulating the Amyloid-$\beta$ protein pathway in AD. It is interesting to observe that both the cortical (precentral gyrus) and subcortical (pallidum) regions responsible for voluntary motor movements \cite{freund2002mechanisms,banker2019neuroanatomy} are all highly responding to AD, indicating a possible link between the behavior and pathological aspect of AD.

\begin{figure}
    \centering
    \includegraphics[width=8cm]{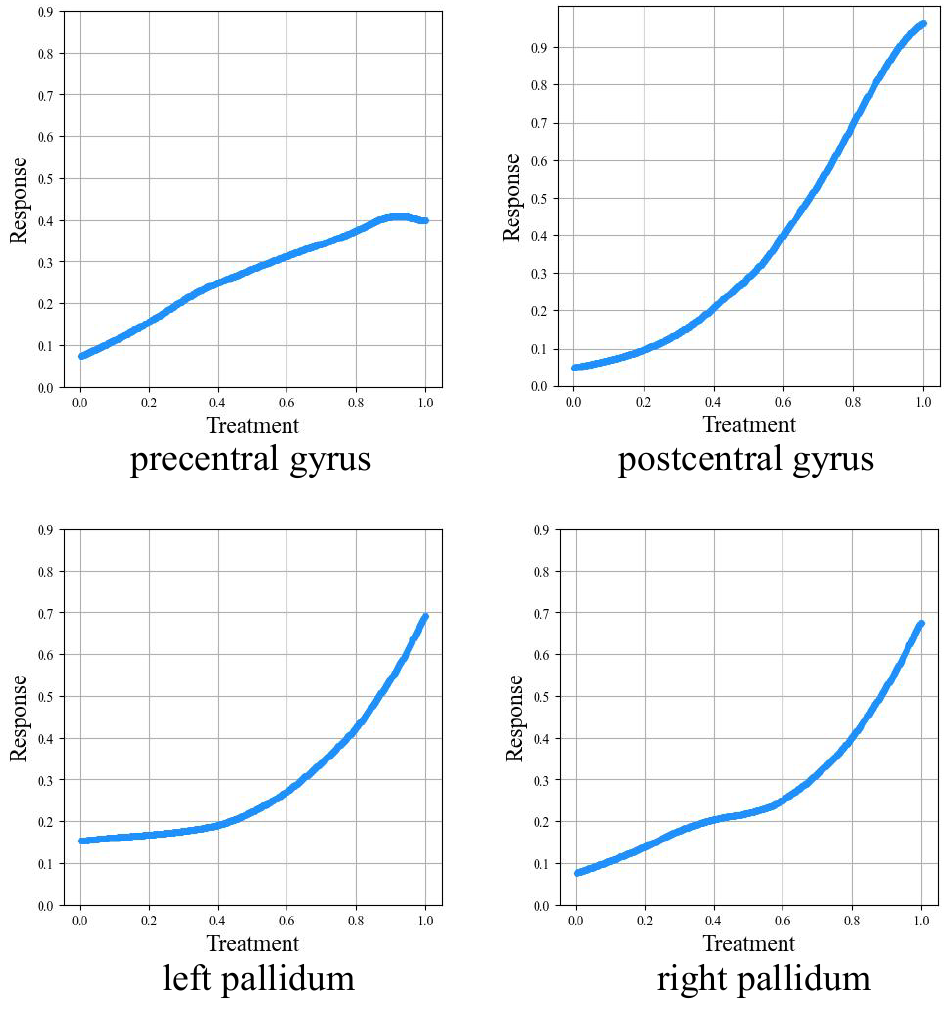}
    \caption{ADRF for the typical upward ROIs.}
    \label{fig:ADRF}
\end{figure}

\begin{figure}[!h]
    \centering
    \includegraphics[width=8cm]{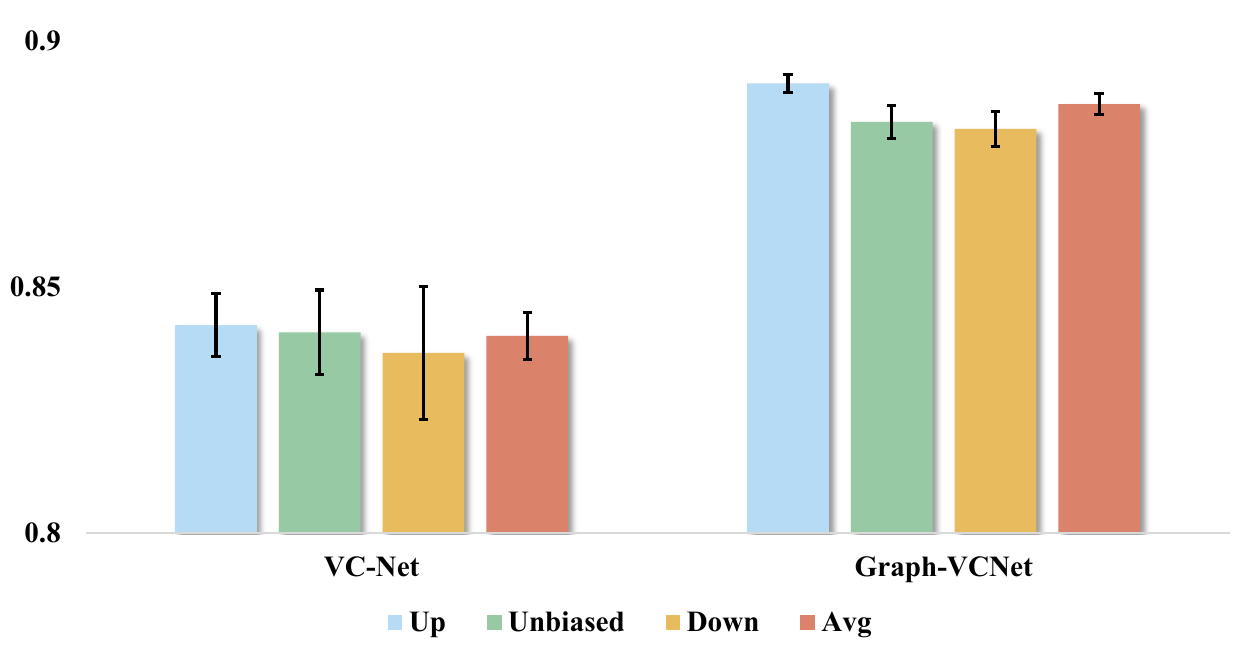}
    \caption{Prediction accuracy with VCNet and Graph-VCNet based on different brain regions.}
    \label{fig:accruacy}
\end{figure}

In addition, based on Table \ref{Tab:cluster} that brain regions in the up group will have a slightly higher prediction power towards the AD probability, we investigated the patterns of ADRF curves and the regions within the up group in Fig. \ref{fig:ADRF}, which is consistent with Figs.  \ref{fig:cort} and  \ref{fig:sub} that pre- and post- central gyrus, left and right pallidum are upward with the increasing treatment.  Moreover, we can obtain the same conclusion from both the VCNet and GVCNet, as shown in Fig. \ref{fig:accruacy}. Compared with the VCNet, our proposed Graph-VCnet can achieve much better prediction accuracy no matter with which kind of brain regions. And more specifically, with upward brain regions, both VCNet and Graph-VCNet could achieve the best prediction accuracy, compared with the other kinds of brain regions.

\begin{table}[h]

\centering
\caption{KMeans Cluster Accuracy}

\begin{tabular}{ll}
\\ \hline
Cluster             & Accuracy \\ \hline
Down    & 0.8836 $\pm$ 0.0034   \\
Unbiased     & 0.8822 $\pm$ 0.0035   \\
Up    & \textbf{0.8915 $\pm$ 0.0018}  \\ \hline
\label{Tab:cluster}
\end{tabular}
\end{table}

\section{Conclusion and Discussion}
In this paper, we propose a novel model called GVCNet, which combines a graph neural network architecture with a targeted regularization approach to estimate varying coefficients of a treatment effect model and improve the model's performance. Experiment results show that GVCNet exhibits promising capabilities in making counterfactual causal inferences for Alzheimer's Disease (AD) progression based on the regional level of Amyloid-beta protein. 

The rationalization for employing a graph neural network architecture in GVCNet stems from the inherent complexity and interconnectedness of brain regions, both structurally, functionally, and pathologically.  The graph structure allows for capturing the potentially long-distance spatial relationships and dependencies among these regions, providing a more comprehensive representation of the underlying proteinopathy dynamics. Furthermore, GVCNet incorporates a targeted regularization approach. Regularization techniques play a crucial role in mitigating model complexity and ensuring robustness. By imposing the proposed regularization constraints, GVCNet can effectively handle the inherent noise and variability in PET imaging data, leading to more reliable, generalizable, and accurate predictions. 

The potential of GVCNet in patient management, treatment, and drug discovery is substantial. If the model demonstrates sufficient robustness and consistency through rigorous validation studies, it can be ultimately utilized to project personalized AD progression trajectories. By leveraging counterfactual analysis, GVCNet can provide insights into the "what if" scenarios by assessing how the current imaging results would evolve if they were to worsen (due to disease progression) or improve (because of the medications or other types of interventions). This information is invaluable in guiding clinicians and patients in making informed decisions about treatment strategies and long-term care plans. Moreover, GVCNet's ability to predict the personalized treatment effect of a patient after administering a medication targeting Amyloid-beta deposition is of significant clinical importance. It can provide insights into the expected outcomes and help determine the optimal dosage for individual patients. This personalized, regional treatment prediction can aid in tailoring interventions and optimizing therapeutic strategies, leading to improved patient outcomes and more efficient use of resources.

Looking ahead, the future of imaging-guided diagnosis, prognosis, and treatment planning for AD is likely to focus on unraveling the underlying mechanisms that link imaging targets, such as Amyloid-beta protein, with the patient’s internal and external characteristics (e.g., genetic factors, health conditions, comorbidities, and social determinants of health) to the disease progression. The proposed counterfactual causal inference modeling approach with multi-modal data input, as demonstrated by GVCNet, will play a pivotal role in this pursuit. With more data modalities and holistic patient characterization, we can uncover critical insights into the disease's pathophysiology, identify novel therapeutic targets, and develop more effective interventions.

In conclusion, counterfactual causal inference modeling such as GVCNet holds immense potential for advancing our understanding of personalized AD management. It will enable personalized projections of disease trajectories and treatment effects, empowering clinicians and patients to make informed decisions. The integration of imaging-guided diagnosis, prognosis, and mechanistic insights will shape the future of AD research and pave the way for improved patient care and therapeutic strategies.

\section*{Declaration of Competing Interest}
The authors declare that they have no known competing financial interests or personal relationships that could have appeared to
influence the work reported in this paper.

\section*{CRediT authorship contribution statement}
\textbf{Haixing Dai} Conceptualization, Formal analysis, Methodology, Software, Writing – original draft. 
\textbf{Mengxuan Hu:} Formal analysis, Methodology. 
\textbf{Qing Li:} Writing – darft \&  review \& editing.
\textbf{Lu Zhang:} Writing – review \& editing.
\textbf{Lin Zhao:} Writing – review \& editing.
\textbf{Dajiang Zhu:} Writing – review \& editing.
\textbf{Ibai Diez:} Writing – review \& editing.
\textbf{Jorge Sepulcre:} Writing – review \& editing.
\textbf{Xingyu Gao:} PET imaging and non-imaging data analysis, writing – review \& editing.
\textbf{Manhua Liu:} Writing – review \& editing.
\textbf{Quanzheng Li:} Writing – review \& editing.
\textbf{Sheng Li:} Writing – review \& editing.
\textbf{Fan Zhang:} Diffusion imaging data analysis and tractography, writing – review \& editing.
\textbf{Tianming Liu:} Conceptualization, Writing – review \& editing. 
\textbf{Xiang Li:} Conceptualization, Writing – review \& editing.

\section*{Acknowledgments}
Data collection and sharing for this project was funded by the Alzheimer's Disease Neuroimaging Initiative (ADNI) (National Institutes of Health Grant U01 AG024904) and DOD ADNI (Department of Defense award number W81XWH-12-2-0012). ADNI is funded by the National Institute on Aging, the National Institute of Biomedical Imaging and Bioengineering, and through generous contributions from the following: Alzheimer's Association; Alzheimer's Drug Discovery Foundation; BioClinica, Inc.; Biogen Idec Inc.; Bristol-Myers Squibb Company; Eisai Inc.; Elan Pharmaceuticals, Inc.; Eli Lilly and Company; F. Hoffmann-La Roche Ltd and its affiliated company Genentech, Inc.; GE Healthcare; Innogenetics, N.V.; IXICO Ltd.; Janssen Alzheimer Immunotherapy Research \& Development, LLC.; Johnson \& Johnson Pharmaceutical Research \& Development LLC.; Medpace, Inc.; Merck \& Co., Inc.; Meso Scale Diagnostics, LLC.; NeuroRx Research; Novartis Pharmaceuticals Corporation; Pfizer Inc.; Piramal Imaging; Servier; Synarc Inc.; and Takeda Pharmaceutical Company. The Canadian Institutes of Health Research is providing funds to support ADNI clinical sites in Canada. Private sector contributions are facilitated by the Foundation for the National Institutes of Health (www.fnih.org). The grantee organization is the Northern California Institute for Research and Education, and the study is coordinated by the Alzheimer's Disease Cooperative Study at the University of California, San Diego. ADNI data are disseminated by the Laboratory for Neuro Imaging at the University of Southern California.

\ifCLASSOPTIONcaptionsoff
  \newpage
\fi

\bibliographystyle{IEEEtran.bst}
\bibliography{mybib}

\end{document}